\documentclass{article}

\usepackage[nonatbib,final]{include/nips_2016_mod_for_arxiv}

\usepackage[utf8]{inputenc} 
\usepackage[T1]{fontenc}    
\usepackage{url}            
\usepackage{booktabs}       
\usepackage{amsfonts}       
\usepackage{nicefrac}       
\usepackage{microtype}      

\usepackage{pgfplots}

\usepackage[utf8]{inputenc}
\usepackage{lmodern}
\usepackage[english]{babel}
\usepackage{amsmath,amssymb,amsthm}
\usepackage{tikz}
\usetikzlibrary{arrows}
\usepackage[numbers]{natbib}

\usepackage{caption}
\usepackage{subcaption}

\newcommand{\nameref}[1]{Section \ref{#1}}

\newcommand{\stress}{\textbf}

\newcommand{\defi}{\emph}

\newtheorem{aProposition}{Proposition}

\newtheorem{aExample}{Example}

\newtheorem{aRemark}{Remark}

\newenvironment{Example}{\vspace{5.5pt}\begin{aExample}}{\end{aExample}}
\newenvironment{Proposition}{\vspace{5.5pt}\begin{aProposition}}{\end{aProposition}}
\newenvironment{Remark}{\vspace{5.5pt}\begin{aRemark}}{\end{aRemark}}

\DeclareMathOperator{\PA}{PA}

\DeclareMathOperator{\pand}{\mathrm{AND}}
\DeclareMathOperator{\pxor}{\mathrm{XOR}}

\newcommand\ind{\protect\mathpalette{\protect\independenT}{\perp}}
\def\independenT#1#2{\mathrel{\rlap{$#1#2$}\mkern2mu{#1#2}}}

\newcommand{\dc}{\mathrm{do}\,}

\newcommand{\se}{\mathrm{H}}
\newcommand{\mi}{\mathrm{I}}
\newcommand{\kl}{\mathrm{D}}

\newcommand{\pa}{\mathrm{pa}}

\newcommand{\vn}[1]{\mathit{#1}} 

\definecolor{light-gray}{gray}{0.75}

\usetikzlibrary{arrows,shapes,plotmarks}
\tikzset{>=stealth'} 
\tikzstyle{graphnode} = [circle,draw=black,minimum size=22pt,text centered,text width=22pt,inner sep=0pt] 
\tikzstyle{var}   =[graphnode,fill=white]
\tikzstyle{const}   =[graphnode,fill=white,draw=none]
\tikzstyle{hid}   =[graphnode,fill=none,draw=gray,text=black,dashed]
\tikzstyle{phantom}   =[graphnode,fill=white,draw=none,text=white]
\tikzstyle{halfhid}   =[graphnode,fill=light-gray,draw=black,text=white]
\tikzstyle{obs}   =[graphnode,fill=none,draw=none]
\tikzstyle{sel}   =[rectangle,fill=none,draw=black]
\tikzstyle{selection}   =[rectangle,draw=none,fill=black,minimum size=5pt,inner sep=0pt,outer sep=0pt]
\tikzstyle{fac}   =[rectangle,draw=black,fill=black!25,minimum size=5pt]
\tikzstyle{facprior} =[rectangle,draw=black,fill=black,text=white,minimum size=5pt]
\tikzstyle{edge}  =[draw=white,double=black,thick,-]
\tikzstyle{prior} =[rectangle, draw=black, fill=black, minimum size=5pt, inner sep=0pt]
\tikzstyle{dirprior} = [circle, draw=black, fill=black, minimum size=5pt, inner sep=0pt]

\newlength\figureheight
\newlength\figurewidth

\definecolor{myred}{rgb}{0.5, 0.0, 0.0}
\definecolor{mynormalred}{named}{red} 
\definecolor{mysinglered}{rgb}{1, 0, 0.0}
\definecolor{mydarkblue}{rgb}{0.0, 0, 0.5}
\definecolor{myblue}{named}{blue} 

\title{Causal inference for data-driven debugging and decision making in cloud computing}

\author{ {\bf Philipp Geiger}\\
Max Planck Institute for \\
Intelligent Systems,\\
T\"ubingen, Germany\\
pgeiger@tuebingen.mpg.de\\
\And
{\bf Lucian Carata}\\
Univeristy of Cambridge,\\
Cambridge, United Kingdom\\
lc525@cam.ac.uk\\
\And
{\bf Bernhard Sch\"olkopf}\\
Max Planck Institute for \\
Intelligent Systems,\\
T\"ubingen, Germany\\
bs@tuebingen.mpg.de
}

\usepackage{parskip}

\begin{document}

\maketitle

\begin{abstract}
Cloud computing involves complex technical and economical systems and interactions. This brings about various challenges, two of which are:
(1) debugging and control to optimize the performance of computing systems, with the help of sandbox experiments, and (2) privacy-preserving prediction of the cost of ``spot'' resources for decision making of cloud clients.

In this paper, we formalize debugging by counterfactual probabilities and control by post-(soft-)interventional probabilities. We prove that counterfactuals can approximately be calculated from a ``stochastic'' graphical causal model (while they are originally defined only for ``deterministic'' functional causal models), and based on this sketch a data-driven approach to address problem (1).
To address problem (2), we formalize bidding by post-(soft-)interventional probabilities and present a simple mathematical result on approximate integration of ``incomplete'' conditional probability distributions.
We show how this can be used by cloud clients to trade off privacy against predictability of the outcome of their bidding actions in a toy scenario.
We report experiments on simulated and real data.
\end{abstract}


\section{Introduction}
In recent years, the paradigm and business model of cloud computing \citep{armbrust2010view} has become increasingly popular. 
It allows to rent computing resources on-demand, and to use them efficiently by sharing them in a smart way, in particular using auctions to sell unused resources.

Several new challenges arise from the paradigm of cloud computing.
On a technical level, it is a problem to understand, control and debug the involved computing systems up to the size of several data centers, with as much automation as possible, to optimize their performance.
We will go into more detail on this in Section \ref{debugging:sec:deb_prob}.
On an economical level, while auctions for ``spot'' resources help providers to use resources more efficiently, the unpredictability of their prices and performance complicates bidding and buying decisions for clients. 
We will go into more detail on this in Section \ref{debugging:sec:priv_prob}.

In the absence of exact models, it is natural to try to address such problems using data-driven methods \citep{padala2009automated,ostrowski2011diagnosing,snee2015soroban,chiang2014matrix,zheng2009justrunit}.
However, standard machine learning usually applies in settings where the underlying system is invariant, often based on the assumption that samples are i.i.d., and does not make predictions about the effect of interventions, which is important though for debugging, decision making and integration of heterogeneous data.

\subsection{Contributions}

The present paper takes first steps towards addressing challenges of cloud computing using {\em causal} models. 
Inferring causal models from (observational) data is notoriously hard, and convincing applications of causal modeling to real world problems are scarce. The present paper is no exception in that the main focus is conceptual rather than empirical.
Our main contributions are:
\begin{itemize}
\item We present two theoretical results for approximations in causal modeling, Propositions \ref{debugging:prop:approx_cf} and \ref{debugging:prop:approx_transf}, which are of relevance for the subsequent cloud problems and possibly beyond, in Section \ref{debugging:sec:pre}.
It needs to be emphasized that the practicability of these theoretical results remains to be proved.
\item In Section \ref{debugging:sec:deb}, we suggest first steps towards causal models and approximate counterfactuals as a principled, data-driven approach for addressing cloud control and performance debugging problems, integrating sandbox experiments.
\item In Section \ref{debugging:sec:priv}, we use approximate integration of causal knowledge to enable cloud clients to better predict performance and costs, while preserving privacy, in a toy setting.
\end{itemize}

\subsection{Structure}

The remainder of this paper is structured as follows:
\begin{itemize}
\item in Section \ref{debugging:sec:pre}, we give brief introductions to causal models and cloud computing;
\item Section \ref{debugging:sec:pre} contains the definition of counterfactuals (in addition to our two theoretical results);
\item Section \ref{debugging:sec:exp} contains simplistic real-world and simulated experiments for our two approaches, as well as a preliminary causal model of a more realistic cloud system;
\item in Section \ref{debugging:sec:related}, we discuss related work;
\item and we conclude the paper with Section \ref{debugging:sec:conclusion}.
\end{itemize}

\section{Background}

\subsection{Causal models}

Here we only give succinct definitions.
For a more detailed introduction to causal models we refer the reader to \citep{Pearl2000,Pearl2009,Spirtes2000,peters2017elements}.

In this paper, we generally assume variables to be discrete, although some results may also hold for the continuous case. 
Let $V$ be a set of variables.
A \defi{graphical causal model (GCM)} \citep{Pearl2000,Spirtes2000} over $V$ consists of
\begin{itemize}
\item  a directed acyclic graph (DAG) $G$ with $V$ as node set, called \defi{causal diagram} or \defi{causal DAG},
\item a conditional probability density $p_{X|\vn{PA}_X=pa_X}$ (for all $pa_X$ in the domain of $\vn{PA}_X$) for each $X \in V$, where $\vn{PA}_X$ are the parents of $X$ in $G$.
\end{itemize}

  %
A \defi{functional causal model (FCM)} $M$ is a special GCM that includes, for each observed variable $X$,
\begin{itemize}
\item a hidden root (i.e., parentless) \defi{background variable} $U_X$ with an arrow only to $X$,
\item such that $X=f_X(PA_X, U_X)$ for some function $f_X$, i.e., $X$ is fully determined by $\PA_X, U_X$.
\end{itemize}
Each FCM \defi{induces} a CGM by dropping the background variables.
By \defi{causal models} we refer to FCMs as well as CGMs.

Given a causal model $M$ and a tuple of variables $Z$ of $M$, the \emph{post-interventional causal model} $M_{\dc Z=z}$ is defined as follows:
drop the variables in $Z$ and all incoming arrows from the causal diagram, and fix the value of variables in $Z$ to the corresponding entry of $z$ in all remaining conditional densities.
Based on this, we define the \defi{post-interventional density of $Y$ after setting $Z$ to $z$ (relative to $M$)}, denoted by $p_{Y|\dc Z=z}$, by the the density of $Y$ in $M_{\dc Z=z}$.
Note that we will use expressions like $p(x|y)$ as shorthand for $p_{X|Y}(x|y)$.

On a non-mathematical level, we consider $M$ to be a correct causal model of some part of reality, if it correctly predicts the outcomes of interventions in that part of reality (clearly there are other reasonable definitions of causation).

\subsection{Cloud computing}

Traditionally, both businesses and individuals have used dedicated local computers, or computer networks, for storing, managing and processing data. 
However, this can be inefficient in several ways: the overhead of maintaining such an infrastructure is high, and one needs to buy enough computers to handle peak loads, while during normal operation most will remain unutilized

Cloud computing significantly changes this, by allowing computing resources to be rented on demand. 
A company, the \defi{cloud provider}, is now responsible for running all the hardware, keeping it upgraded and sharing it amongst multiple clients. 
Such an infrastructure can be run in a highly efficient manner: tens or hundreds of \emph{virtual machines (VMs)}, i.e., emulations of computer systems, chartered by different clients, run on a single \emph{physical server} and share its resources such as central processing units (CPUs), memory and network.
Note that we refer to a system as being \defi{in production} if this system does actual work for clients and visitors and if contracts have to be met w.r.t.\ this system (in contrast, e.g., to an experimental system).


\section{Two approximations in causal modeling}
\label{debugging:sec:pre}
\label{debugging:sec:defi_causal_model}
\label{debugging:sec:sel_dia}
\label{debugging:sec:rl}

\subsection{Structural counterfactuals and an approximation}
\label{debugging:sec:structural_cf}
\label{debugging:sec:approx_cf}

Let $M_0$ be an FCM over a set $V$ of variables, and let $U$ denote the set of independent background variables in $M_0$. 
Let $E, X, Y$ be (sets of) variables in $V$.
The \defi{structural counterfactual probability of $Y$ being $y$, had $X$ been $x$, given evidence $E=e$}, can be defined \citep{Pearl2000} based on $M_0$ as\footnote{Note that \citep{Pearl2000} in his definition uses functions instead of (deterministic) conditionals.}
\begin{align}
p( Y_{\dc X=x}=y | e) := \sum_u p( y | \dc(x), u) p(u | e) . \label{debugging:eqn:scf}
\end{align}



Even though computer systems are ``more deterministic'' than many other systems, due to interactions with the environment and missing information, usually one can only infer a GCM, and not an FCM, of a computer system.
Without an FCM though, counterfactual probabilities (Equation (\ref{debugging:eqn:scf})) are generally not uniquely determined, i.e., they cannot be derived from a GCM.
Let us give an example.

\begin{Example}[GCMs do not determine counterfactual probabilities]
\label{debugging:expl:cf}
Let $V= \{X,Y\}$ for binary $X,Y$, and consider the GCM $M$ with DAG $X \to Y$ and conditionals $p_X(0) = \frac{1}{2}$ and $p_{Y|X}(0|x) = p_Y(0) = \frac{1}{2}$.
$M$ is induced by two very different FCMs.
On the one hand, the FCM $M_0$ with structural equations
\begin{align*}
X := U_X, \\
Y := U_Y,
\end{align*}
and $U_X \sim U_Y \sim Uniform(\{0,1\})$, where $Uniform(\{0,1\})$ denotes the uniform distribution on $\{0,1\}$, induces $M$.
On the other hand, the FCM $M_0'$ with structural equations
\begin{align*}
X := U_X, \\
Y := X \pxor U_Y,
\end{align*}
and $U_X \sim U_Y \sim Uniform(\{0,1\})$ induces $M$.
But in $M_0$ we have
\begin{align*}
&p( Y_{\dc X=1}=0 | X=0, Y=0) \\
&= \sum_{u_Y} p( Y=0 | \dc X=1, u_Y) p(u_Y | X=0, Y=0) \\
&= \sum_{u_Y} p( Y=0 | u_Y) p(u_Y | X=0, Y=0) \\
&= 1 \cdot 1 + 0 \cdot 0 = 1,
\end{align*}
while in $M_0'$ we have
\begin{align*}
&p( Y_{\dc X=1}=0 | X=0, Y=0) \\
&= \sum_{u_Y} p( Y=0 | \dc X=1, u_Y) p(u_Y | X=0, Y=0) \\
&= \sum_{u_Y} p( Y=0 | X=1, u_Y) p(u_Y | X=0, Y=0) \\
&= 0 \cdot 1 + 1 \cdot 0 = 0.
\end{align*}
This gives an extreme example of counterfactual  probabilities not being determined by a GCM.
\end{Example}
For a more detailed discussion of this phenomenon we refer the reader to \citep{peters2017elements}. 

Now we show that nonetheless counterfactual probabilities can be calculated \emph{approximately}, and one can \emph{know}, from only the GCM, how wrong the approximation is at most -- on average.
This will be important for our approach to debugging in Section \ref{debugging:sec:deb}, and, as we belief, for other areas as well.

Let $M$ be a GCM and let $Z$ be the set of its root variables (variables with no parents in the causal DAG). 
For any (sets of) variables $X,Y,E$ in $M$ we define the \emph{approximate structural counterfactual} or \emph{approximate counterfactual} as%
\footnote{The idea of a counterfactual definition based on only the GCM has been mentioned in \citep[Section 7.2.2]{Pearl2000}, but not been further investigated. 
Depending on the specific setting and the available information, there may be more suitable approximations to encode counterfactual-like probabilities.
}
\begin{align}
\tilde{p}( Y_{\dc X=x}=y | e )  := \sum_w p( y | \dc(x), w ) p( w | e ) , \label{debugging:eqn:approx_cf}
\end{align}
where $W := Z \setminus X$.

\begin{Proposition}
\label{debugging:prop:approx_cf}
Let $M_0$ be an FCM that induces a GCM $M$, and let $Z$ denote the root variables in $M$.
For all (sets of) variables $E, X, Y$ we have
\begin{align}
\kl( p( Y_{\dc X=x} | E ) \| \tilde{p}( Y_{\dc X=x} | E ) ) \leq \se(E | Z ) ,\label{debugging:eqn:acf_prop}
\end{align}
where $p( Y_{\dc X=x} | e )$ is defined w.r.t.\ $M_0$ and $\tilde{p}( Y_{\dc X=x} | e )$ w.r.t.\ $M$.
\end{Proposition}
We prove (using monotonicity of the KL divergence and properties of entropy) a generalization of Proposition \ref{debugging:prop:approx_cf} -- Proposition \ref{debugging:prop:approx_cf_gen} -- in Section \ref{debugging:sec:prop_cf}.\footnote{Note that, if we chose the set $Z$ in Proposition \ref{debugging:prop:approx_cf_gen} such that it is as ``close'' (in the causal diagram) to $Y$ as possible, this could yield better approximations than simply letting $Z$ be the root nodes, as done in $\bar{p}( Y_{\dc X=x} = y| e )$.
We leave this as a question for future work.}

\begin{Example}
To give some intuition about the approximate counterfactual and the proposition, let us first consider the following two special cases:
If $M$ is already an ``FCM'' in the sense that all its variables are completely determined by the root nodes, then we have $H(E | Z) = 0$, and thus, based on Equation (\ref{debugging:eqn:acf_prop}), both quantities coincide, which seems natural.
If the evidence comprises the root nodes, $Z \subset E$, then the approximation amounts to the simple conditional $p( y | \dc (x), w )$ (where $w$ is the part of $e$ the corresponds to $W$), similar as if we had evidence on all background variables in an FCM.

Note that for the $M$ in Example \ref{debugging:expl:cf}, the approximate counterfactual does not help much.
It can be calculated as
\begin{align*}
\tilde{p}( Y_{\dc X=1}=0 | X=0, Y=0) = p( Y=0 | \dc(x) ) = p(Y=0) = \frac{1}{2} 
\end{align*}
As is easy to see, this implies the KL divergence between $\tilde{p}( Y_{\dc X=1} | X=0, Y=0)$ and the true $p( Y_{\dc X=1} | X=0, Y=0)$ under both, $M_0$ and $M_0'$ of Example \ref{debugging:expl:cf}, to be 1.
This KL divergence coincides with the upper bound to the KL divergence in Proposition \ref{debugging:prop:approx_cf}, since $H(X,Y | Y)=1$ in Example \ref{debugging:expl:cf}.
\end{Example}

The practical meaningfulness of the approximate counterfactual probability, in particular for decision making, remains subject to debate.
We will briefly comment on it in Remark \ref{debugging:rem:cf} below.

\subsection{Approximate integration of causal knowledge}
\label{debugging:sec:approx_transf}

The following result will be important for Section \ref{debugging:sec:priv} since it can be used to preserve some amount of \emph{privacy}.
Consider random variables $C, X_0,\ldots,X_K, Z$. 
A typical causal structure which satisfies the assumptions we make below is depicted in Figure \ref{debugging:fig:priv_example} on page \pageref{debugging:fig:priv_example}. 
Here we introduce what can be seen as an approximation to ``transportability'', as introduced by \citet{Pearl2011,bareinboim2012transportability} (with various subsequent extensions and versions \cite{bareinboim2013general,bareinboim2014transportability,GeiHofSch16}), in the following simple case: we would like to know $p(z)$, we do know the mechanism $p(z|x_0, \ldots, x_K)$ plus, from a different source, $p(x_k,c)$ for all $k$, but we do \emph{not} know $p(x_0, \ldots, x_K)$.
Define the approximation
\begin{align}
\label{debugging:eqn:apint}
 \bar{p}(z) := \sum_{x_0,\ldots,x_K,c}  p(z|x_0,\ldots,x_K) \prod_k p(x_k|c) p(c) .
\end{align}

\begin{Proposition}
\label{debugging:prop:approx_transf}
If $Z \ind C | X_0, \ldots, X_K$, then
$
\kl( p(Z) \| \bar{p}(Z) ) \leq \sum_k \se(X_k |C ) .
$
\end{Proposition}
Note that based on the proposition, again, we can \emph{know} how wrong the approximation is at most, using only the available information $p(x_k|c),p(c)$.
A proof (again using monotonicity of the KL divergence and properties of entropy), can be found in Section \ref{debugging:sec:prop_transf}.

\begin{Example}
To get an intuition, consider the case that all $X_k$ are fully determined by $C$: then $\bar{p}(z)$ and $p(z)$ coincide, which is reflected by $\sum_k \se(X_k |C )$ being 0.
As already mentioned, an example of a causal model which implies the condition of the proposition is depicted in Figure \ref{debugging:fig:priv_example} on page \pageref{debugging:fig:priv_example}.
\end{Example}

While here we apply the proposition for a predictability-privacy problem in Section \ref{debugging:sec:priv}, it is more generally applicable where joint distributions are not available.
In particular, while in Section \ref{debugging:sec:priv} we will focus on approximate integration for \emph{privacy} reasons, an even more frequent reason may be that only (insufficient) marginals are \emph{known}.
Keep in mind that stronger statements on the set of possible $p(z)$ under the available information may exist, e.g., based on ideas in \cite{balke1994counterfactual}.

\section{Problem 1 -- models for control and debugging -- and our approach}
\label{debugging:sec:deb}

We start with the problem statement (Section \ref{debugging:sec:deb_prob}), followed by our approach (Section \ref{debugging:sec:deb_approach}).
Then we illustrate our approach in detail based on several toy scenarios and discuss advantages over previous work (Section \ref{debugging:sec:deb_disc}).

\subsection{Problem statement}
\label{debugging:sec:deb_prob}

Cloud computing involves technical systems of the highest complexity, which have to be controlled and debugged, ideally in a (semi-)automatic way.
More specifically, the \defi{control problem} can be stated as follows:
During the operation of a cloud server many decisions automatically have to be made regarding how resources, such as complete computers, or parts, such as CPU time, are allocated among the various applications or virtual machines (VMs) of clients.
The goal is to optimize this automatic decision making, based on some given utility function, encoding, e.g., energy consumption, guarantees given to customers, or simply profit.

The \defi{(performance) debugging problem} (closely related to ``performance attribution'') can be formulated as follows:
the general goal is to understand which component of a system contributes to what extent to the measured performance. Based on this, it can be decided which components have to be modified to perform a ``gradient step'' towards the optimal performance. To give an example, a cloud computing client may wonder whether the high latency of his web server is caused from concurrent programs within his VM (which he could directly intervene on), or by other, concurrent VMs on the same physical cloud sever. We will come back to this example in Section \ref{debugging:sec:deb_disc}, where we address a toy scenario, as well as Section \ref{debugging:sec:realistic}, where we give an example of a preliminary but realistic causal model that can help in such a situation.  
Note that we presently focus on debugging for \emph{individual observations}, i.e., on the unit-level. 

Usually, plenty of heterogeneous knowledge and data is available about the involved systems:
expert knowledge, formal program code and system specifications (often containing non-causal associational knowledge), data from the very system or similar ones, and data from sandbox experiments.

\subsection{Outline of our approach}
\label{debugging:sec:deb_approach}

We now sketch several steps of a unified approach based on causal models, which can potentially help to address the control and the debugging problem.
In what follows, we will refer to the cloud system ``in production'', i.e., the fully configured system with a specific set of applications, as the \defi{``target system''}.
Note that, depending on the specific setup, some steps may be canceled.

\subsubsection[Step 1A]{Step 1A: inference of causal diagram and some mechanisms}
\label{debugging:sec:deb_step_inf}

\emph{Given:} 
the various information sources described below.

\emph{Procedure:}
Keep in mind that the inference procedure we describe here is usually not based on the target system itself, since some details of it (such as the specific VMs running on it) are varying quickly, but instead on \emph{past} experience with other systems of equal or similar configuration.
In particular, usually not all details of the target system are known during this step, so that some mechanisms stay underdetermined, but can be inferred later during \nameref{debugging:sec:deb_step_exp}.
As usual, the main sources for causal inference are randomized interventional experiments, observational data, and expert knowledge. 
A necessary condition to harness the first two sources is the decision about - and performance of - \emph{measurements} of the system, for which we propose to use tools discussed by \citet{Carata2014,snee2015soroban}.

Note the important fact that many aspects of computer systems (hardware and software) are - \emph{by design} - modular, i.e., separable into individually manipulable input-output mechanisms, which is a central assumption in causal models \cite{Pearl2000,peters2017elements}.
To give a simple example: to see if erroneous behavior is caused by the network, one can unplug the network cable and check if the error occurs nonetheless -- a procedure which generally would not change any other mechanism, such as the CPU or keyboard.
Furthermore, the same (or similar) mechanisms occur in different systems, which is very helpful for extrapolation from experiments.
Note that there is an additional source of information which is specific to computer systems: a lot of knowledge about non-causal associations, such as which program calls which other program during execution, is available, often in a well-formatted way (e.g. program code or system architecture specifications). Such information could be translated into hypotheses on causal association (or be used for measurement selection), in a (semi-)automatic way.

The output of this procedure is a causal diagram $G$ of the target system, together with those mechanisms, i.e., conditionals in the causal model $M$ of the target system, which can be inferred based on past experience.
For those mechanisms which cannot be known based on past experience, but only when the target system is revealed (e.g., the specific VMs running on it), but which \emph{cannot be explored directly on the target system} either (since tentative configurations may violate contracts with clients \citep{chiang2014matrix,zheng2009justrunit}), we discuss the integration of sandbox experiments in \nameref{debugging:sec:deb_step_exp} below, which should then complete the causal model $M$.

\subsubsection[Step 1B]{Step 1B: design and integration of sandbox experiments}
\label{debugging:sec:deb_step_exp}

\emph{Given:} 
an additional cloud system, the \defi{``experimental system''}, equivalent in hardware to the target system, the causal diagram $G$ of the target system, some variable $X$ (e.g. performance of some VM) in $G$, and the identity (e.g., VM) but not all properties of the mechanism that produces $X$, and whose unknown properties should be inferred during the experiment.

\emph{Procedure:}
The knowledge of $G$ allows to integrate sandbox experiments in a principled way: 
\begin{enumerate}
\item Derive all direct influences of $X$ from $G$, i.e., the parents $\PA_X$ (which could include resources such as CPU time or size of requests received from the internet).
\item \emph{Design} the sandbox experiment on the experimental system such that (1) the experimental system has the same conditional $p(x|\pa_X)$ as mechanism for $X$ (e.g., by simply running the same VM on the experimental system as is planned to run on the target system) and (2) all variables in $\PA_X$ are randomly varied.
\item Based on the gathered data, regress $X$ on $\PA_X$ and plug the inferred conditional $p(x|\vn{pa}_X) = p(x|\dc \vn{pa}_X)$ as mechanism for $X$ into $M$.
This is possible since all parents of $X$ were ``intervened'' and regressed upon.
\end{enumerate}

Without going further into detail, it needs to be mentioned that the transfer of the conditional between experimental and target system can be seen as a simple example of ``transportation'' of causal relations as defined by \citet{pearl2011transportability}.


\subsubsection[Step 1C]{Step 1C: control}
\label{debugging:sec:deb_step_control}

\emph{Given:}
causal model $M$ of the target system, 
some utility $U$, which is variable in $M$ or a function of one or several variables in $M$, 
and some variable $X$ (e.g. concurrent workload, CPU time, network bandwidth) in $M$, which should be controlled such as to optimize $u$ (or $p(u)$).

\emph{Procedure:}
As $M$ predicts the effect on $u$ of modifying any of its mechanisms, it can be used to find the mechanism, or ``policy'', $p(x|\pa_X)=\pi(x|pa_X)$, which maximizes $u$.

\subsubsection[Step 1D]{Step 1D: observation-level performance debugging}
\label{debugging:sec:deb_step_deb}

\emph{Given:}
causal model $M$ of the target system, 
a variable $Y$ in $M$ that measures the performance which we want to optimize,
a performance debugging query $Q$,
and an (individual) observation $Y=y,F=f$, where $F$ contains all observables besides $Y$. (Since we move on the level of individual observations instead of populations, we term this step ``observation-level performance debugging''.)

\emph{Procedure:}
For the performance debugging query $Q$, we assume the following form:
``In the current situation, would it improve performance $Y$ from the current $y$ to $y'$, if we would set $X$ to $x'$, given side information $F=f$?''
The side information $f$ may contain an observation $x$ of $X$.
Stated this way, it seems natural to translate this query into a query for the structural counterfactual probability $p(Y_{\dc X=x'}=y' | y, f)$.\footnote{Clearly, there are other ways to formalize attribution and debugging.}
%
Then, based on Section \ref{debugging:sec:approx_cf} and in particular Proposition \ref{debugging:prop:approx_cf}, we can calculate the approximate answer $\tilde{p}(Y_{\dc X=x'}=y' | y, f)$ from the GCM $M$, if $\se(E|Z)$ is small, where $Z$ is a set of root nodes.


\begin{Remark}[The value of (approximate) counterfactuals for performance debugging]
\label{debugging:rem:cf}
A remark is due regarding the notion of a counterfactual and its application to performance debugging.
In the narrow sense, a counterfactual statement is always a statement about the past and so it is neither falsifiable, nor can it help for any (falsifiable recommendations regarding) future decision.

In contrast, here we have in mind a broader notion of a counterfactual: a situation where one observes a system with a poor performance and asks how the performance could be debugged when the system remains in the \emph{``same''} state, or visits the same or similar states again. (In the language of causal models, ``state'' means the tuple of background variables.)
This question is relevant in situations where the debugging action can be performed quickly after the observation of poor performance, and where ones assumes that the \emph{state changes comparably slowly}, i.e., the state varies smoothly with time.\footnote{It seems that a more thorough analysis of this argument might be fruitful, as it could theoretically justify the frequent usage of counterfactual reasoning in everyday life.
We leave this to future work.}
Alternatively, the question can be relevant if one has a good ``subjective'' judgement about the similarity of the state between two points in time -- if the judgement is based on pbjective observables though, a non-counterfactual form of reasoning may be more appropriate.

Situations where counterfactual reasoning may be useful arise, in particular, whenever one does not assume to ``know'' the population-level distribution of the state well enough (but one beliefs in the structural equations), for instance, because it varies with time, and instead one wants to reason on the observation-level, i.e., unit-level. Because on the population-level, there are better ways for decision making than counterfactual reasoning, see \nameref{debugging:sec:deb_step_control}.

We propose one way to formalize performance debugging questions, and to answer them, based on one possible formalization of counterfactual probabilities proposed by \citet{Pearl2000}.
It remains an open question whether there are better formalizations than ours for the debugging questions we consider, and whether the general notion of a counterfactual probability, as well as its formalization by \citet{Pearl2000}, are sensible. For a discussion, see also \citep{peters2017elements}.

Note that an additional issue, which we are not able to settle here, is how close our approximation of a counterfactual comes to the true counterfactual \emph{in practice}.
\end{Remark}

\subsection{Application to toy scenarios and discussion of potential advantages over previous approaches} 
\label{debugging:sec:deb_disc}

For researchers familiar with causal inference, some of the steps described above may seem trivial.
However, all current approaches to the described problems we are aware of are lacking a \emph{principled} (formal) language, with concepts such as causal sufficiency, for such things as integration of sandbox experiments and performance debugging.

We will now give toy examples to make the approach outlined in \nameref{debugging:sec:deb_step_exp} through \nameref{debugging:sec:deb_step_deb} more concrete, and simultaneously show the advantages of our approach based on causal models over some previous approaches. (For examples of applications of \nameref{debugging:sec:priv_step_pick}, see Sections \ref{debugging:sec:deb_experiment} and \ref{debugging:sec:exp_realisic}.)
Keep in mind that, clearly, the approach we outlined does not completely solve the problem: the inference of knowledge it relies on remains a challenge as with all other approaches. However, our approach may be less prone to errors and more data-efficient.

\begin{itemize}
\item \stress{\nameref{debugging:sec:deb_step_exp}:}
Integrating sandbox experiments without a principled approach  \citep{chiang2014matrix,zheng2009justrunit}, can lead to errors:
e.g., if not all parents (direct causes) of a variable $X$ are varied during the experiment and regressed upon afterwards or, say, $X$ is regressed on its causal children.
Any methodology that does not include reasoning about concepts such as causal effect, causal sufficiency or randomization is prone to such mistakes.
Let us give a toy example of how our approach works for sandbox experiments, and how other approaches can go wrong in terms of variation and regression. 

\begin{Example}[Design and integration of sandbox experiments, and possible mistakes]
\label{debugging:expl:sand}

Imagine we are the cloud provider and we want to decide whether we can put some VM $A$ on some cloud server, where already other concurrent VMs are running. Let $L \in \{0,1\}$ denote the performance of (the main application running inside) $A$, with $L=0$ denoting good, and $L=1$ bad performance. For instance, $L$ could denote some latency.
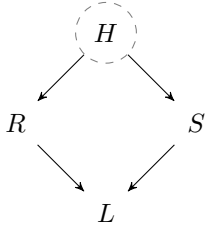
\begin{figure}
	\centering 
	\begin{tikzpicture}[scale=0.6]
	
	\node[obs] at (0, 0) (P) {$L$};
	
	\node[obs] at (-2, 2) (R) {$R$} edge[->] (P);
	\node[obs] at (2, 2) (S) {$S$} edge[->] (P);
	
	\node[hid] at (0, 4) (T) {$H$} edge[->] (R) edge[->] (S);
	
	
	\end{tikzpicture}
	\caption{Causal diagram when running VM $A$ on the target system. Not varying $S$ or not regressing on $S$ during the sandbox experiment can lead to wrong predictions of performance $L$ on the target system, especially when some hidden source $H$ (say internet users) introduces strong correlations between $R$ and $S$.} 
	\label{debugging:fig:deb_adv}
\end{figure}
Assume that Figure \ref{debugging:fig:deb_adv} depicts the correct causal DAG of the target system, i.e., when $A$ would be running on the mentioned cloud sever. In particular, the performance depends on two factors, say amount of requests $R \in \{0,1\}$ coming into $A$ from the internet, on the one hand, and usage $S \in \{0,1\}$ of the CPU of the cloud sever by the concurrent VMs, on the other, where 0 stands for ``low'' and 1 for ``high''.
And in turn, $R,S$ depend on $H$ which may denote the state of the internet users, which send requests to $A$ but potentially also to concurrent VMs and therefore also influence $S$.
(Alternatively, $H$ could denote a \emph{parameter} for the behaviour of the internet users, i.e., for the distribution of their states.)

Assume the true mechanism underlying $L$ to be 
\[L := R \pand S, \]
where $\pand$ denotes the logical ``AND''.
I.e., the performance is bad iff $A$ has to serve many requests ($R=1$) and at the same time CPU usage by concurrent VMs is high ($S=1$).
Furthermore, assume that on the target system, we have $R\approx S$.
For instance, this could be due to the fact that $A$ and concurrent VMs serve internet users in the same time zone. Additionally, assume $p(R=0)=p(S=0)=\frac{1}{2}$.

Suppose we have inferred the causal DAG in Figure \ref{debugging:fig:deb_adv} based on \nameref{debugging:sec:deb_step_inf}.
We now want to infer the mechanism underlying the performance $L$, so, following \nameref{debugging:sec:deb_step_exp} (taking $L$ as $X$), we would perform a sandbox experiment where we would vary both, $R$ and $S$, and afterwards regress on both, $R$ and $S$. We would correctly infer the mechanism $L := R \pand S$. Additionally knowing $p(r,s)$ (say from previous experience, or from reports by the cloud clients) we would correctly predict the probability of bad performance of $A$ on the target system, $p(L=1)$, to be
\begin{align*}
\sum_{r,s} p(L=1|r,s) p(r,s) = 0 \cdot \frac{1}{2} + 1 \cdot \frac{1}{2} = \frac{1}{2}
\end{align*}

In contrast, without such a principled approach, two things can happen.

If in the sandbox experiment, only $R$ is varied and regressed upon, while $S$ is kept to a constant $0$ (because it was not properly inferred or communicated as an influence factor, or simply because on the experimental system no concurrent VMs are emulated), then $p(L=1)$ would be wrongly predicted as 
\begin{align*}
\sum_{r} p(L=1|r,0) p(r) = 0 \cdot \frac{1}{2} + 0 \cdot \frac{1}{2} = 0.
\end{align*}

And even if in the sandbox experiment, $S$ would be varied according to the correct $p(s)$ on the target system (e.g., because the concurrent VMs of the target system would be emulated well on the experimental system), but if one would \emph{forget about regressing on $S$}, then still one would wrongly predict  $p(L=1)$ to be
\begin{align*}
\sum_{r} p(L=1|r,s) p(r) p(s) = 0 \cdot \frac{1}{4} + 0 \cdot \frac{1}{4} + 0 \cdot \frac{1}{4} + 1 \cdot \frac{1}{4} = \frac{1}{4}.
\end{align*}
	
Clearly, this was only a simplistic toy example, but to the best of the knowledge of the author, such problems have not been thematized in the literature \citep{chiang2014matrix,zheng2009justrunit} yet.
\end{Example}

\item \stress{\nameref{debugging:sec:deb_step_control}:}
Causal models provide a principled tool for control of cloud systems that allows to integrate various forms of information, such as results of sandbox experiments obtained in \ref{debugging:sec:deb_step_exp}.
Furthermore, compared to, e.g., \citep{padala2009automated}, which is based on adaptive control, an advantage of using causal models is that they allow to encode and integrate knowledge about which mechanisms vary and which stay invariant. 

\begin{Example}[Control based on causal models]
\label{debugging:expl:control}

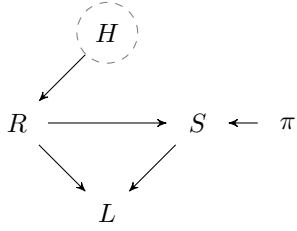
\begin{figure}
\centering 
\begin{tikzpicture}[scale=0.6]

\node[obs] at (0, 0) (P) {$L$};

\node[obs] at (2, 2) (S) {$S$} edge[->] (P);
\node[obs] at (-2, 2) (R) {$R$} edge[->] (P) edge[->] (S);

\node[hid] at (0, 4) (T) {$H$} edge[->] (R);

\node[obs] at (4, 2) (pi) {$\pi$} edge[->] (S);

\end{tikzpicture}
\caption{Causal diagram when running $A$ on a system controlled by policy $\pi$.
It is similar to the system in Figure \ref{debugging:fig:deb_adv}, except that now $S$ is influenced by the choice of the policy (as well as the current $R$ which serves as an input to the policy), and therefore we add $\pi$ to the diagram and draw an arrow to $S$. Note that handling the policy, which is rather a parameter than a variable, in such a way is similar to the use of so-called ``selection diagrams'' in \cite{pearl2011transportability}, where the mechanisms that vary are marked by special nodes with arrows to them.} 
\label{debugging:fig:control_adv}
\end{figure}

Consider $A$, the same VM as in Example \ref{debugging:expl:sand}, with performance $L$.
Recall that there we inferred the mechanism for $L$ to be $L := R \pand S$.

Now assume that we consider a different target system than in Example \ref{debugging:expl:sand}, namely, a system that involves a policy $\pi(r|s)$ that controls the amount $S$ of CPU that is occupied by VMs other than $A$. We depict the causal DAG in Figure \ref{debugging:fig:control_adv}.

Suppose the goal is as follows: keep the probability of poor performance below $\frac{1}{2}$, i.e., $p(L=1|\pi) \leq \frac{1}{2}$, while allocating as little CPU as possible to $A$, i.e., minimizing $p(S=0|\pi)$ (so that more CPU can be used by other VMs).
Furthermore, assume $p(R=0) = \frac{1}{2}$, as in Example \ref{debugging:expl:sand}.

Using the causal DAG and ``plugging in'' our knowledge of the mechanisms, it is easy to see that the optimal policy is $\pi(S=1|r) = 1$, i.e., always occupy the CPU by other VMs. Because then
\begin{align}
p(L=1|\pi) &= \sum_{r,s} p(L=1|r,s) \pi(s|r)  p(r) \label{debugging:sec:vary} \\
&= p(L=1|0,1) \frac{1}{2} + p(L=1|1,1) \frac{1}{2} \\
&= 0 + \frac{1}{2} = \frac{1}{2}
\end{align}
so the goal w.r.t.\ performance $L$ is still met.
This shows how causal models provide a principled tool to integrate sandbox experiments, based on \nameref{debugging:sec:deb_step_exp}, to perform control, as proposed in \nameref{debugging:sec:deb_step_control} (the $S$ here corresponds to the $X$ there).

Let us mention a potential advantage of control based on causal models in case cloud systems are time-varying.
Assume $H$ denotes a parameter for the behavior of the internet users (we indicated this meaning in Example \ref{debugging:sec:deb_step_exp}).
Suppose $H$ varies for some reason, say due to an ad campaign, in an unpredictable way.
We know that the behavior of the internet users influences $L$ only via $R$, since the rest of the cloud system is not affected by the internet.
This knowledge is encoded in the causal DAG in Figure \ref{debugging:fig:control_adv}.
Based on this, we have
\begin{align*}
p(l|r,s,h) = p(l|r,s).
\end{align*}
So we have formally reasoned that even if $H$ varies, the mechanism $p(l|r,s)$ stays the same.
Hence, to derive the new optimal policy $\pi$, all one has to do is to infer the new $p(r)$ and plug it into Equation \ref{debugging:sec:vary} (and optimize for $\pi$).
Furthermore, we can be certain that we \emph{identified} the new system and the new optimal policy (given our assumptions are correct).
This sort of reasoning has been analyzed, on a more general level, by \citet{pearl2011transportability} (but they do not apply it to control settings).

%
%
%
%
%

In contrast, approaches to (adaptive) control for cloud computing which are not based on modularity and such reasoning \citep{padala2009automated} may try to infer the complete information, $p(r)$ as well as $p(l|r,s)$, from scratch upon a variation of $H$, assuming it to be a completely new ``environment''. And even if such approaches utilize the invariant $p(l|r,s)$ after a variation of $H$ in one way or another, they are usually missing the language to \emph{reason about the identifiability} of the new system (after the variation in $H$), as we did above based on causal models.

It needs to be emphasized that here we considered an overly simplistic scenario.
In more complex and realistic scenarios, there are much more mechanisms involved that could potentially vary or stay invariant, respectively.
See Section \ref{debugging:sec:realistic} for an example of a causal DAG of a more realistic but still simple cloud system.

\end{Example}

\item \stress{\nameref{debugging:sec:deb_step_deb}:}
We now give en example for how observation-level debugging can be performed based on \nameref{debugging:sec:deb_step_deb}. This approach can be seen as complementary to other methods for this problem \citep{ostrowski2011diagnosing}, where errors may arise from confusing causation with correlation, or where it is more difficult to integrate heterogeneous knowledge such as sandbox experiments. 

\begin{Example}[Observation-level performance debugging]
\label{debugging:expl:deb}

Note that, while this is a toy scenario, the assumptions we make in this example regarding what is known/observed and what not are close to realistic \citep{snee2015soroban}.

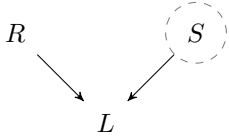
\begin{figure}
	\centering 
	\begin{tikzpicture}[scale=0.6]
	
	\node[obs] at (0, 0) (P) {$L$};
	
	\node[obs] at (-2, 2) (R) {$R$} edge[->] (P);
	\node[hid] at (2, 2) (S) {$S$} edge[->] (P);
	
	
	
	\end{tikzpicture}
	\caption{Causal diagram for observation-level performance debugging in a toy setting. $S$ is unobserved, but nonetheless we assume $p(l|r,s)$ to be known, may it be that the provider publishes it, or the client knows it from own (sandbox) experiments.} 
	\label{debugging:fig:deb_deb}
\end{figure}

Similar as in Example \ref{debugging:expl:sand}, consider a VM with performance (latency) $L$ running on a cloud system, with $R \in \{0,1\}$ denoting the amount of incoming requests, and $S \in \{0,1\}$ the amount of, say, CPU time allocated to concurrent VMs (0 stands for ``low'' and 1 for ``high''). 
Here, denote the VM by $B$.
In contrast to Example \ref{debugging:expl:sand}, assume the causal DAG depicted in Figure \ref{debugging:fig:deb_deb}.
Furthermore, let $L \in \{0,1,2,3\}$ and the structural equation for $L$ be given by
\begin{align}
L := R + S + U_L , \label{debugging:eqn:cfs}
\end{align}
with $p(U_L = 0) = \frac{1}{2}$.
Suppose $p(S=R) = p(S=0) = \frac{1}{2}$, where $p(S=0) = \frac{1}{2}$ may be seen as encoding some prior belief.

Now assume that the client whom $B$ belongs to wonders, whether it would improve the latency $L$ to a desired $0$ in the current situation where she observes $L=2, R=1$, if she decreased the amount of incoming requests to a lower level, i.e., if she set $R$ to $0$.
(Note that ``current situation'' can include the nearby future, if the unobserved variables vary comparably slowly, see Remark \ref{debugging:rem:cf}.)
She does \emph{not} observe $S$ due to neither the cloud provider nor other clients publishing this information.
This is a realistic assumption in cloud computing.
Based on \nameref{debugging:sec:deb_step_deb}, she translates this question into a query for the counterfactual probability $p( L_{\dc R=0}=0 | R=1, L=2)$.

Suppose that while $S$ is not published, $p(l|r,s)$ is known, may it be that the provider publishes it, or the client knows it from own (sandbox) experiments.
That is, $p(r), p(s)$ and $p(l|r,s)$ are give, but not the structural Equation \ref{debugging:eqn:cfs} itself.
Now, although the structural Equation \ref{debugging:eqn:cfs} would be needed to calculate the counterfactual $p( L_{\dc R=0}=0 | R=1, L=2)$ exactly (see Example \ref{debugging:expl:cf}) she can calculate the \emph{approximate} counterfactual probability defined in Equation \ref{debugging:eqn:approx_cf} as
\begin{align*}
&\tilde{p}( L_{\dc R=0}=0 | R=1, L=2) \notag\\
&= \sum_s p(L=0|\dc R=0, s) p(s|R=1, L=2) \notag\\
&= \sum_s p(L=0|\dc R=0, s) p(L=2|R=1, s) p(s|R=1) \frac{1}{p(L=2|R=1)}   \notag\\
&= \frac{1}{4}, 
\end{align*}
where we plugged in $R,S$ for the set of root variables $Z$, which yields $S$ as $W$, and as evidence $E$ we took $(R,L)$ with value $(1,2)$.
Based on this, she concludes that the probability that setting $R$ to $0$ helps for decreasing latency $L$ to $0$ is rather small (in the current situation).

Note that the true counterfactual probability (Equation \ref{debugging:eqn:scf} in Section \ref{debugging:sec:approx_cf}) in this specific case is given by
\begin{align*}
&p( L_{\dc R=0}=0 | R=1, L=2) \\
&= \sum_{u_R, u_S, u_L} p(L=0|\dc R=0, u_R, u_S, u_L) p(u_R, u_S, u_L|R=1, L=2) \\
&= \sum_{u_L, s} p(L=0|\dc R=0, s, u_L) p(s,u_L|R=1, L=2)  \\
&= 0 + p(L=0|\dc R=0, S=0, U_L=1) p(S=0, U_L=1|R=1, L=2) \\
&\phantom{=}+ p(L=0|\dc R=0, S=1, U_L=0) p(S=1, U_L=0|R=1, L=2) + 0  \\
&= 0,
\end{align*}
which would lead to the even stronger conclusion that setting $R$ to $0$ for decreasing $L$ to $0$ would not work at all.

Note that the upper bound of Proposition \ref{debugging:prop:approx_cf} here takes the value
\begin{align*}
\se(R,L|R,S) &= \se(L|R,S) \\
&= \sum_{r,s} p(r,s) \se(L|r,s) \\
&= 1 .
\end{align*}
Recall that we picked $p(U_L = 0)=\frac{1}{2}$, i.e., rather strong noise.
For less noise, the approximation would be even better and the bound smaller.

\end{Example}

\end{itemize}

Note that generally, one could try to \emph{learn} (in the sense of machine learning) things such as how to perform and integrate the experiment \citep{snee2015soroban}, but one would always have to rely on prior assumptions, which may then be more difficult to encode.

%
%
%
%
%
%
%
%
%
%
%
%
%
%
%
%
%
%
%
%

\section{Problem 2 -- cost predictability versus privacy -- and our approach}
\label{debugging:sec:priv}

We start with the problem statement (Section \ref{debugging:sec:priv_prob}), followed by our approach (Section \ref{debugging:sec:priv_approach}).
Then we present a toy example (Section \ref{debugging:sec:priv_expl}), and some additional remarks (Section \ref{debugging:sec:priv_disc}).

\subsection{Problem statement}
\label{debugging:sec:priv_prob}

Here we consider an economical aspect of cloud computing.
Currently, one common way for clients to purchase cloud resources from a provider is via an auction mechanism for ``spot'' (i.e., short-term) resources, which can be described in a simplified way as follows:
The customer enters a bid, e.g., for an hour of usage.
Once the price determined by the provider (based on supply, demand, and other private factors) drops below the bid, the customer gets the resource, usually as long as her bid exceeds the price (within the hour).
This approach has several advantages, in particular for the provider: he can sell resources which are unused but which fluctuate a lot (due to guarantees given to ``dedicated'' or ``on-demand'' customers).
But clients can profit as well: the spot resources are usually significantly cheaper than the long-term dedicated resources.

An obvious drawback of spot resources is that this kind of mechanism comes with a high \emph{uncertainty} for the clients:
it is hard to tell how the prices will evolve in the future, and, in particular, purchased resources can be terminated in an unforeseeable way, which is, to some extent, due to the \emph{unpredictability of the other clients}.
Therefore, if the client does not want to take these risks which can significantly harm his/her business, they often avoid this mechanism.


\subsection{Sketch of our approach}
\label{debugging:sec:priv_approach}

In what follows, we present a very first step towards addressing the problem based on available observational data and causal models.
We assume that there is one provider, and clients $1,\ldots,K$. By ``stakeholders'' we refer to provider and clients together.
For each point in time (say, the beginning of an hour), 
let $X_k$ denote client $k$'s demand for the next hour, 
$Y_k$ the cloud product that the client buys from the provider, 
$W_k$ the information based on which the client decides her demand (e.g., hour of the day), which may not always be fully known though,
and $\pi_k$ her policy determining which cloud product $Y_k$ to buy, given her demand $X_k$.
Let $X_0$ denote the provider's pricing parameter at that time point (which may depend, e.g., on energy costs),
and let $Z$ denote the outcome of the provider's mechanism applied to the $Y_k$.
(Generally, $Z$ can include the price as well as say termination of spot resources; for simplicity, let it only denote the cost/price for the moment, which can comprise the indirect costs resulting from loss of visitors through termination.)

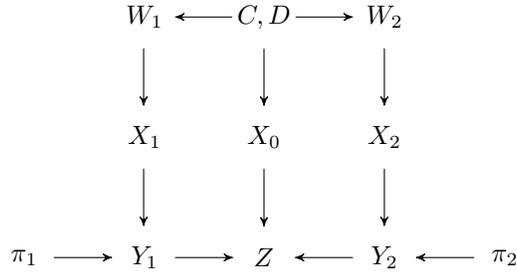
\begin{figure}
\centering 
\begin{tikzpicture}[scale=0.8]
    
\node[obs] at (0, 0) (Z) {$Z$};
\node[obs] at (-2, 0) (Y_1) {$Y_1$} edge[->] (Z);
\node[obs] at (2, 0) (Y_2) {$Y_2$} edge[->] (Z);

\node[obs] at (-4, 0) (pi_1) {$\pi_1$} edge[->] (Y_1);
\node[obs] at (4, 0) (pi_2) {$\pi_2$} edge[->] (Y_2);

\node[obs] at (-2, 2) (X_1) {$X_1$} edge[->] (Y_1);
\node[obs] at (2, 2) (X_2) {$X_2$} edge[->] (Y_2);
\node[obs] at (0, 2) (X_0) {$X_0$} edge[->] (Z);

\node[obs] at (-2, 4) (W_1) {$W_1$} edge[->] (X_1);
\node[obs] at (2, 4) (W_2) {$W_2$} edge[->] (X_2);


\node[obs] at (0, 4) (H) {$C,D$} edge[->] (W_1) edge[->] (W_2) edge[->] (X_0); 


\end{tikzpicture}
\caption{Causal diagram $G_2$. $D$ is hidden.} 
\label{debugging:fig:priv_example}
\end{figure}

We assume the following simple mechanism (which is a simplified version of the auction described above): all clients $k$ always get the product they want, but the subsequent price vector $Z$ varies and is not known in advance.
The causal diagram $G_2$ for the complete causal structure, for the case $K=2$, is depicted in Figure \ref{debugging:fig:priv_example}. The role of $C$ will be explained below, while $D$ denotes the hidden part of the confounder $(C,D)$.


Our approach to the uncertainty problem, towards more predictable prices and subsequent reduced costs, is based on the idea that clients may not want to share all, but are willing to share some of their information between each other.
More specifically, we propose the following two-step procedure which allows the clients to \emph{trade off privacy versus predictability interests}, by jointly picking a variable $C$ such that $p(X_k|C)$ allows an approximate prediction of $Z$ which still preserves some privacy.%
\footnote{An extreme approach would be to directly infer a joint model for all clients from their joint data (i.e., considering all clients as a ``single client'').
Here we assume that this is not possible, due to heterogeneous data, privacy interests, etc.}

\subsubsection[Step 1A]{Step 1A: jointly picking $C$}
\label{debugging:sec:priv_step_pick}

First, all stakeholders $k$ pick their candidates for $C$ (possibly based on a given list and some ``privacy budget''), balancing their privacy interests against minimizing $\se(X_k|C)$.
If the intersection of their candidates is non-empty, they reveal $\se(X_k|C)$ for all $k$ and joint candidates $C$.\footnote{If the intersection is empty, the procedure is canceled without result, and the stakeholders proceed in the classical, non-collaborative way.
}
They pick the $C$ that minimizes $\sum_k \se(X_k|C)$ to optimize the predictability, based on Proposition \ref{debugging:prop:approx_transf}.

\subsubsection[Step 2B]{Step 2B: prediction and individual decision}
\label{debugging:sec:priv_step_pred}

Now all clients $k$ reveal their $p(x_k|c)$. $p(c)$ is assumed to be common knowledge.
Furthermore, all $p(y_k|x_k,\pi_k)$
are either known a priori (based on the possible products the provider offers) or revealed now.
The provider reveals $p(z|x_0,y_0,\ldots,y_K)$ and $p(x_0|c)$.
Now $\bar{p}(z|\pi_1,\ldots,\pi_K)$ can be calculated, based on Equation \ref{debugging:eqn:apint}.
More specifically, we have
\begin{align*}
\bar{p}(z|\pi_1,\ldots,\pi_K) &= \sum_{x_0,\ldots,x_K}  p(z|x_0,\ldots,x_K;\pi_1,\ldots,\pi_K) \prod_{k=0}^K p(x_k|c) p(c) \\
&=\sum_{x_0,\ldots,x_K}  \left( \sum_{y_1,\ldots,y_K} p(z|x_0,y_1,\ldots,y_K) \prod_{k=1}^K p(y_k|x_k;\pi_k) \right) \prod_{k=0}^K p(x_k|c) p(c)
\end{align*}

Then, based on Proposition \ref{debugging:prop:approx_transf}, the clients narrow down the set of possible $p(z|\pi_1,\ldots,\pi_K)$ to those for which  \[ \kl(p(Z|\pi_1,\ldots,\pi_K) \| \bar{p}(Z|\pi_1,\ldots,\pi_K)) \leq \sum_k \se(X_k|C). \]
Based on this constraint on $p(z|\pi_1,\ldots,\pi_K)$, each client $k$ decides on their strategy $\pi_k$, e.g., based on game-theoretic considerations.

\subsection{Application to toy scenario}
\label{debugging:sec:priv_expl}

To illustrate the approach, let us give an example.

\begin{Example}
\label{debugging:expl:priv}
A cloud provider, Clark, offers to his clients, Alice ($k=1$) and Bob ($k=2$), monthly (dedicated) large resources ($Y_k=2$), rather expensive, or hourly spot small ($Y_k=0$) and large ($Y_k=1$) resources, which are usually cheaper.
However, if Alice and Bob happen to both order large spot resource for the same hour, the cost for both of them ($[Z]_1, [Z]_2$) is significantly higher than the hourly rate for the monthly large resource, since Clark may have to buy a new resource, or he may have to cancel one of his client's applications, causing the loss of web site visitors.
Now assume Alice and Bob, during \nameref{debugging:sec:priv_step_pick}, pick the hourly weather forecast, which is $0$ for sunny and $1$ for cloudy, for $C$, since it is public information anyway that both their web sites are  \emph{weather related}: Alice runs a website for \emph{outdoor} activities, Bob one for \emph{indoor} activities, both in the same region.
And the remaining uncertainty w.r.t. their demand ($X_k$ being 0 for ``small'' or 1 for ``high''), i.e., $H(X_k|C)$, is small.
The causal diagram for this scenario is $G_2$ depicted in Figure \ref{debugging:fig:priv_example}.
Based on this, Alice and Bob can conclude that they will rarely require a large resource at the same time, and they can go for spot resources as their respective (dominant) strategies $\pi_k$.
\end{Example}

\subsection{Discussion}
\label{debugging:sec:priv_disc}

In some cases, the provider could infer the joint distribution of all $X_k$, based on past data, which would contain all relevant information.
However, the complete system is so complex that it is unlikely to be \emph{stationary}.
Note that during each step, already some information is revealed, but this is transparent to the stakeholders.
Limitations of our approach are that (1) the clients may not even be willing to reveal their $p(x_k)$, or (2) $X_k$ may not be predictable or the model may be wrong (although humans and organizations usually do plan ahead).

It needs to be emphasized, that here we ignore strategic aspects, which can lead to problems in certain scenarios. As a potential next step, such aspects could be analyzed based on game theory \citep{shoham2008multiagent}.

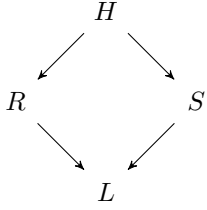
\begin{figure}
\centering 
\begin{tikzpicture}[scale=0.6]
    
\node[obs] at (0, 0) (P) {$L$};

\node[obs] at (-2, 2) (R) {$R$} edge[->] (P);
\node[obs] at (2, 2) (S) {$S$} edge[->] (P);

\node[obs] at (0, 4) (T) {$H$} edge[->] (R) edge[->] (S);


\end{tikzpicture}
\caption{Causal diagram $G_1$.} 
\label{debugging:fig:deb_example}
\end{figure}

%
%
%
%
%
%
%
%
%
%
%
%
%
%
%
%
%
%
%
%
%
%
%
%
%
%
%
%
%
%

\section{Experiments}
\label{debugging:sec:exp}


\subsection{Control and debugging problem on simple but real cloud system}
\label{debugging:sec:deb_experiment}

Here we test small parts of our approach in Section \ref{debugging:sec:deb_approach} on a very simple, but real cloud system: a physical server running a specific application (a web server) together with some concurrent workload (another web server).
The system we consider has the same causal DAG as two of the examples in Section \ref{debugging:sec:deb_disc}. And while the scenarios are generally similar, the system we consider here is simpler, for experimental purposes.

A source $H$ keeps sending simultaneous request to application and concurrent workload (drawn from a multivariate correlated Poisson distribution), of which $R$ are received by the application and $S$ by the concurrent workload.
Then, for each request, the latency (performance) of the application is measured in nanoseconds by $L$.


\begin{figure}
\centering
\setlength{\figurewidth}{0.5\textwidth}
\setlength{\figureheight}{0.35\textwidth}
\input{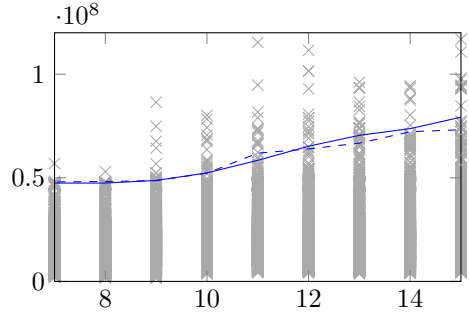}

\caption{
X-axis: Number of simultaneous requests $S=s$. Y-axis: 99th percentile of prediction $\hat{p}(l|\dc s)$ (dashed blue) is close to 99th percentile (solid blue) of ground truth test data from $p(l|\dc s)$ (subsample in gray).}
\label{debugging:fig:plot_deb}
\end{figure}

We examine how well \nameref{debugging:sec:deb_step_inf} works.
First, we infer the causal diagram $G_1$ depicted in Figure \ref{debugging:fig:deb_example}, as well as an estimate of $p(r,s,l)$ from observational samples of the system, based on \nameref{debugging:sec:deb_step_inf}, and together denote them by (incomplete) $M_1$.
Then, from $M_1$, using back-door adjustment \cite{Pearl2000}, we derive a prediction $\hat{p}(l|\dc s)$ for $p(l|\dc s)$.
Besides \nameref{debugging:sec:deb_step_inf}, this tests the applicability of \nameref{debugging:sec:deb_step_control}, when thinking of a simple controller that outputs a constant for $S$ (e.g. by putting the application on another machine with such a concurrent workload), as well as \nameref{debugging:sec:deb_step_deb} which relies on post-interventional distributions (of an updated model, though).
The outcome is depicted in Figure \ref{debugging:fig:plot_deb}, where we use the 99th percentile as statistic, which is common in cloud computing.
It shows that the prediction is close to the ground truth test data, both in magnitude and in trend.

\subsection{Example of a more realistic cloud system}
\label{debugging:sec:realistic}
\label{debugging:sec:exp_realisic}

The experiments in the previous Section \ref{debugging:sec:deb_experiment} were performed on an overly simplistic system. Here we want to give an example of a preliminary, partial causal model (causal DAG plus some knowledge on the mechanisms, e.g., additivity) of a more \emph{realistic} system to which our approach in Section \ref{debugging:sec:deb_approach}, in particular the performance debugging in \nameref{debugging:sec:deb_step_deb}, is meant to be applied. Note that this is merely for illustration purposes, we do \emph{not test} any hypothesis here.\footnote{The inference of the causal model of -- and the application of our approach to -- such a complex system turned out to be more difficult than expected. Therefore, no evaluation of our approach applied to this system can be reported at this stage.}

We consider a cloud sever running serveral VMs.
We focus on one specific VM, call it $A$ for the moment.
Inside the VM $A$, a web sever $B$ (more specifically: ``lighttpd'') runs. We consider the following observed and hidden variables, among others, measured inside and outside $A$\footnote{In cloud computing, it is important to distinguish between inside and outside of $A$, since, for privacy reasons, often only things inside $A$ can be known to the client that $A$ belongs to.}:
\begin{itemize}
\item ``req\_size'': size of the file requested by an internet user from the web server $B$;
\item ``local\_load'': resource-consuming activity of other applications in $A$, besides $B$;
\item ``concurrent\_vm\_count'': number of VMs running concurrently with $A$ on the physical sever (outside $A$);
\item ``srv\_lat'': latency of the web server $B$, which can be seen as part of the \emph{objective} which needs to be minimized.
\end{itemize}
\begin{figure}
	\centering 
	\includegraphics[width=\textwidth]{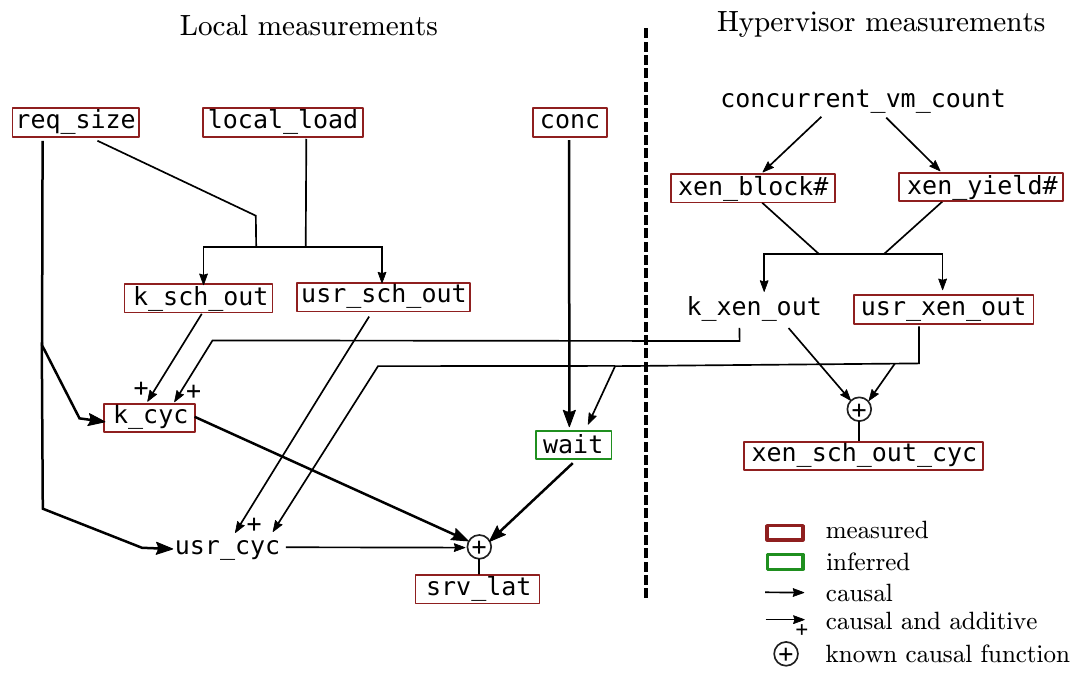}
	\caption{Example of a preliminary causal DAG of a cloud system. Variables on the left side are measured within the VM $A$, that runs together with other VMs on the cloud server. The right side contains measurements outside the VM (the ``hypervisor'' is the program that is responsible for allocating the cloud server's resources among the VMs). 
		The objective is to minimize the latency of some web server $B$ running in $A$, denoted by ``srv\_lat'', while keeping utilization by other VMs, denoted by ``concurrent\_vm\_count'', as high as possible. Possible \emph{manipulations} include reducing the workload within the VM $A$, denoted by ``local\_load'', versus changing the number of concurrent VMs. If the causal model is good, it can help to pick the optimal manipulations. The figure is taken from \citep{carata2016} which also gives descriptions of the remaining variables not described here.}
	\label{debugging:fig:realistic}
\end{figure}
We depict the partial causal model in Figure \ref{debugging:fig:realistic}.
It is taken from \citet{carata2016}, who also gives descriptions for all other variables in the figure not discussed here.
Note that this is a model of an \emph{experimental} system, while on a system in production, some variables, such as  ``local\_load'' and ``concurrent\_vm\_count'', could be influences by a (hidden) common cause, similar to $H$ in the previous experiment in Section \ref{debugging:sec:deb_experiment}.

This model was inferred as described in \nameref{debugging:sec:deb_step_inf}, in an iterative and sequential way, based on non-causal associational knowledge about the program execution structure (known from the program code) as well as the general system architecture, further expert knowledge, and independence tests on sampled data. As can be seen, often the integrated knowledge allows to draw conclusions on the \emph{additivity} of mechanisms, which can be based on the fact the runtime of one program essentially is the sum of the runtimes of its subroutines.

Such a model could help for decision making in various ways, for instance for the performance debugging problem mentioned in Sections \ref{debugging:sec:deb_prob} and \ref{debugging:sec:deb_disc}:
A cloud client, the owner of $A$, may observe a high latency (``srv\_lat'') of his web server $B$, together with some other variables.
He wonders if, in this situation, the high latency is caused by other programs within his VM $A$ (``local\_load''), by other, concurrent VMs (``concurrent\_vm\_count'') running on the same physical cloud sever, or simply by large requests (``req\_size'') coming in at that moment.
Based on this, he could conclude whether he should intervene on ``local\_load'', which may be the simplest, or rather intervene on ``concurrent\_vm\_count'' say by changing to another cloud product, such as a dedicated server, which may be more expensive.

%

\subsection{Predictability-privacy problem on simulated data}

For our approach in Section \ref{debugging:sec:priv_approach} to work, $\bar{p}(z)$ has to approximate $p(z)$ reasonably well.
Here we examine to what extent this is the case in a simulated version of the toy example in Section \ref{debugging:sec:priv_expl}, additionally testing how tight the bound in Proposition \ref{debugging:prop:approx_transf} is.
Compared to the toy example, we restrict ourselves to spot resources, i.e., $\{0,1\}$ for $Y_k$, and assume the following specific mechanisms:
The policy $\pi_k$ is for both to simply purchase their demand ($Y_k := X_k$), Clark's pricing is ``cheap'' ($Z=0$) versus ``very expensive for one of them since both want large'' ($Z=1$), in particular $Z:= Y_1 \pand Y_2$.
Furthermore, \[X_k := C \pxor D \pxor N_{X_k},\] where $D$ is some confounder which Alice and Bob do not want to reveal.

\begin{figure}
	\centering
	\setlength{\figurewidth}{0.5\textwidth}
	\setlength{\figureheight}{0.35\textwidth}
%
%
%
\begin{tikzpicture}

\begin{axis}[
xmin=0, xmax=0.5,
ymin=0, ymax=0.5,
axis on top,
width=\figurewidth,
height=\figureheight
]
\addplot [blue]
table {%
0 0.165
0.01 0.158
0.02 0.166
0.03 0.158
0.04 0.133
0.05 0.149
0.06 0.157
0.07 0.15
0.08 0.144
0.09 0.156
0.1 0.155
0.11 0.152
0.12 0.135
0.13 0.157
0.14 0.159
0.15 0.127
0.16 0.16
0.17 0.139
0.18 0.144
0.19 0.129
0.2 0.131
0.21 0.123
0.22 0.126
0.23 0.151
0.24 0.122
0.25 0.118
0.26 0.128
0.27 0.126
0.28 0.127
0.29 0.113
0.3 0.105
0.31 0.117
0.32 0.125
0.33 0.112
0.34 0.116
0.35 0.105
0.36 0.124
0.37 0.116
0.38 0.112
0.39 0.093
0.4 0.108
0.41 0.118
0.42 0.091
0.43 0.101
0.44 0.103
0.45 0.103
0.46 0.091
0.47 0.113
0.48 0.092
0.49 0.106
0.5 0.097
};
\addplot [blue, dashed]
table {%
0 0.165222159856309
0.01 0.169416890681004
0.02 0.16907363928289
0.03 0.166877294189858
0.04 0.1530649563582
0.05 0.164040780141844
0.06 0.167988219911135
0.07 0.180802025630426
0.08 0.168502420865978
0.09 0.190572358054186
0.1 0.184734166666667
0.11 0.181966063395753
0.12 0.17578925604158
0.13 0.208656832298137
0.14 0.212932703321879
0.15 0.191371951219512
0.16 0.209126544656126
0.17 0.201712803365333
0.18 0.201161150864639
0.19 0.203579992030209
0.2 0.210699789325843
0.21 0.205131912734652
0.22 0.206864357180147
0.23 0.227014724064291
0.24 0.218406756642051
0.25 0.213338254000134
0.26 0.227923839164915
0.27 0.223392586674078
0.28 0.229687712954593
0.29 0.231996420460016
0.3 0.220780777667503
0.31 0.229807228915663
0.32 0.225486040202508
0.33 0.228407486524399
0.34 0.230138181818182
0.35 0.244816616008105
0.36 0.245548692498568
0.37 0.246288636363636
0.38 0.242763157182672
0.39 0.235871925851283
0.4 0.240289651115435
0.41 0.247637003849641
0.42 0.24365
0.43 0.242619047619048
0.44 0.254718654738897
0.45 0.25660063559322
0.46 0.238220602172487
0.47 0.25844045174538
0.48 0.24611092078718
0.49 0.249993902439024
0.5 0.243033
};
\addplot [red]
table {%
0 6.45554221107882e-08
0.01 0.000170145039495209
0.02 1.21860586206929e-05
0.03 0.000103712211791738
0.04 0.000581593821976361
0.05 0.000305180759217689
0.06 0.00015859476809604
0.07 0.00121145954977534
0.08 0.000805110704900202
0.09 0.00146900488020303
0.1 0.00110674416643125
0.11 0.00113882929267345
0.12 0.00221483127606528
0.13 0.00311961980624598
0.14 0.00335436524361167
0.15 0.0053360640504614
0.16 0.00280574387392064
0.17 0.0048166121033191
0.18 0.00397496398565403
0.19 0.00689319640286045
0.2 0.00769627632466418
0.21 0.00841287947381329
0.22 0.00807361434975426
0.23 0.00651647944048598
0.24 0.0112085878401919
0.25 0.0111867706634226
0.26 0.011640466240618
0.27 0.0112188807975052
0.28 0.0122611310728685
0.29 0.016767043548
0.3 0.0165724077983194
0.31 0.0150435108335904
0.32 0.0119010124402181
0.33 0.0162168441327961
0.34 0.015414862509374
0.35 0.0227728787991844
0.36 0.016639434468988
0.37 0.0193370313084157
0.38 0.0197695244868195
0.39 0.0248547727692206
0.4 0.0204944481872538
0.41 0.0190231410466524
0.42 0.027947843896933
0.43 0.0236565967812619
0.44 0.026310943735582
0.45 0.0268554709788738
0.46 0.0263424370348769
0.47 0.0235772598161287
0.48 0.0282623185593964
0.49 0.023826296216896
0.5 0.0253188097243665
};
\addplot [red, dashed]
table {%
0 0.360543703510664
0.01 0.364751576840714
0.02 0.363147580560558
0.03 0.369260517141857
0.04 0.370208274756959
0.05 0.381171299419019
0.06 0.374562384747367
0.07 0.39332654867113
0.08 0.39023589884458
0.09 0.40523318220316
0.1 0.390559165419679
0.11 0.403045467042017
0.12 0.409629258323712
0.13 0.417981855656832
0.14 0.427834898907016
0.15 0.421709016302767
0.16 0.417720343789602
0.17 0.425908231506804
0.18 0.425709800093808
0.19 0.430064879692915
0.2 0.451907434774146
0.21 0.429526317462004
0.22 0.441759452973999
0.23 0.445334376204459
0.24 0.456365304991618
0.25 0.450844469801783
0.26 0.468425691847657
0.27 0.457002854285556
0.28 0.468772716671535
0.29 0.461012367607346
0.3 0.471586844407032
0.31 0.480719879231755
0.32 0.465001297036948
0.33 0.473892631967081
0.34 0.47841750115321
0.35 0.482667445913001
0.36 0.48766487767916
0.37 0.486867905823799
0.38 0.489999053760154
0.39 0.491634611234849
0.4 0.484310812081957
0.41 0.493394437302868
0.42 0.495820747201082
0.43 0.491061002021817
0.44 0.497798959002687
0.45 0.499176846447638
0.46 0.496511953504678
0.47 0.497388278132552
0.48 0.499509027476188
0.49 0.499388572863836
0.5 0.499906217741193
};
\end{axis}

\end{tikzpicture}

\captionof{figure}{X-axis: Parameter $r$ (the higher, the more influence from $D$). Y-axis: $\bar{p}(z)$ (dashed blue) is close to $p(z)$ (solid blue) even when $D$ gets strong and $C$ weakens; $\frac{1}{4} \kl(p(Z)\|\bar{p}(Z))$ (solid red), $\frac{1}{4} \sum_k \se(X_x|C)$ (dashed red).}
\label{debugging:fig:plot_priv}
\end{figure}
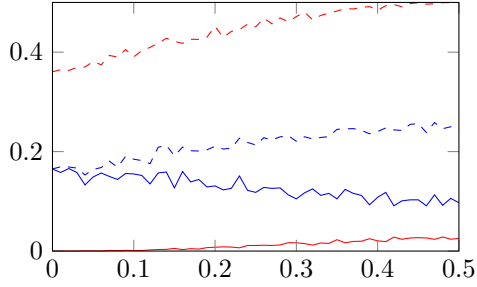

Now for ``each'' $0 \leq r \leq 0.5$, we draw 1000 samples of $C \sim \vn{Bernoulli}(0.5-r)$, $D \sim \vn{Bernoulli}(r)$ to find out how wrong $\bar{p}(z)$ gets when increasing the confounder $D$ that is not revealed or adjusted for, and $N_{X_k} \sim \vn{Bernoulli}(0.2 - 0.2 r)$ (to also examine a little variation in the noise strength).
The outcome is depicted in Figure \ref{debugging:fig:plot_priv}.
It shows that $\bar{p}(z)$ is a good estimate in this simple setting (which is also due to the fact that already $p(x_1), p(x_2)$ alone reveal something about $p(x_1, x_2)$).
It also shows that (in this setting), the bound from Proposition \ref{debugging:prop:approx_transf} may be improvable, as the dashed red line is far away from the solid red line.

\section{Related work}
\label{debugging:sec:related}

Regarding Section \ref{debugging:sec:pre}, approximations to non-identifiable quantities in causal models were examined by \citet{balke1994counterfactual}.
While their technique does not seem directly applicable to the setup of Proposition \ref{debugging:prop:approx_cf}, it may allow to derive stronger statements, i.e., further narrowing down the set of possible $p(z)$, than Proposition \ref{debugging:prop:approx_transf}, which could be examined in future work.
We discussed some related work for Section \ref{debugging:sec:deb}, i.e., the control and debugging problem, in Section \ref{debugging:sec:deb_disc}.
Additionally, maybe the work closest to our investigation in that section is by \citet{lemeire2007causal}, which suggests to use causal models for performance modeling of programs, but does not consider counterfactuals, or more complex computing systems.
Generally, the utilization of modularity based on causal models in that section is strongly inspired by the theory of ``transportability'' of causal relations developed by \citet{pearl2011transportability}, however, that theory has not been applied to (cloud) computing problems so far.
The relation between causality and control is also considered by \citet{bottou2013counterfactual}.
Regarding Section \ref{debugging:sec:priv}, the work by \citet{angel2014end} can be seen as related in that they allow the provider to hide their exact costs while still making some information of the costs available to others.
The work by \citet{mcsherry2007mechanism} investigates privacy-preserving mechanisms, but does not consider the integration of the revealed information to an (estimate) of a causal model.

\section{Conclusions}
\label{debugging:sec:conclusion}

This paper assayed how causal inference can, in principle, help with technological and economical problems in cloud computing.
Guided by these problems, we presented two theoretical results for approximate causal inference, and reported initial experimental results. The application of causal inference in this domain is, to the best of our knowledge, the first of its kind. We believe the potential in this area is very significant, both for applications and for methodological work. Problems in computing systems, which often require sophisticated interventions to bring the system closer to its ``optimum'', rarely fit the classical settings that machine learning excels at.\footnote{Note that cloud computing systems are a good example where application of data-driven methods has to be understood and optimized within highly complex socio-technical interactions and not just in isolation as was the classical focus.}

In particular, for issues such as integration of sandbox experiments, (formally) reasoning about concepts such as causation, causal sufficiency and randomization seems crucial, and methodology which neglects this, such as classical machine learning, may be prone to errors. Another concept which plays an important role in causal modeling (but, of course, also in some other areas) is that of identifiability, which helps to ``critically'' reason about what can and what cannot be inferred based on the given. We used it for the control problem for cases that only some ``modules'' of the system vary.

A next step would be to extend the experimtens on real cloud systems, such as the system for which a preliminary model was derived in Section \ref{debugging:sec:realistic}, and based on this, advance the approach we sketched in Section \ref{debugging:sec:deb}. Another future step would be to use aspects of game theory and mechanism design, to extend our approach for the predictability-privacy trade-off in Section \ref{debugging:sec:priv}.

\appendix

\section{Appendix}

Here we present proofs for Section \ref{debugging:sec:pre}.

\subsection{Generalized version and proof of Proposition \ref{debugging:prop:approx_cf}}
\label{debugging:sec:prop_cf}

We start by stating and proving a generalization of Proposition \ref{debugging:prop:approx_cf}.

\begin{Proposition}[Generalization of Proposition \ref{debugging:prop:approx_cf}]
\label{debugging:prop:approx_cf_gen}
Let $M_0$ be a FCM over discrete variables that induces a GCM $M$.
Let the triple $(X, Y, Z)$ of (sets of) variables in $M$ be such that $(Y \ind An(Z) | Z)_{M}$ (i.e., are d-separated \citep{Pearl2000}) and $X$ does not influence $W := Z \setminus X$.
Let $E$ be an arbitrary set of variables in $M$. 
Let 
\begin{align}
p^Z( Y_{\dc X=x}=y | e )  := \sum_w p( y | \dc X=x, w ) p( w | e ) . \label{debugging:eqn:approx_cf_gen}
\end{align}
Then
\begin{align}
\kl( p( Y_{\dc X=x} | E ) \| p^Z( Y_{\dc X=x} | E ) ) \leq \se(E | Z ) 
\end{align}
(where $p( Y_{\dc X=x} | E )$ is defined w.r.t.\ $M_0$ and $p^Z( Y_{\dc X=x} | E )$ w.r.t.\ $M$).
\end{Proposition}

This is a generalization of Proposition \ref{debugging:prop:approx_cf}.
To see this, let $Z$ denote the set of root nodes of $M$.
This implies 
\[ 
p^Z( Y_{\dc X=x}=y | e ) = \tilde{p}( Y_{\dc X=x}=y | e )
\]
for $p^Z( Y_{\dc X=x}=y | e )$ as defined above and $\tilde{p}( Y_{\dc X=x}=y | e )$ as defined as in Section \ref{debugging:sec:approx_cf}.
But Proposition \ref{debugging:prop:approx_cf_gen} above applied to this $p^Z( Y_{\dc X=x}=y | e )$ coincides with Proposition \ref{debugging:prop:approx_cf}.


\begin{proof}
Let $U_1$ be the set (tuple) of background variables that influence $Z$ and $U_0 = U \setminus U_1$.
Then
\begin{align}
&p^Z( Y_{\dc X=x} = y| e ) \\
&= \sum_w p( y | \dc X=x, w ) p( w | e ) \\
&= \sum_{w,u_0} p( y | \dc X=x, w, u_0 ) p( w | e ) p(u_0 | \dc X=x, w) \\
&= \sum_{w,u_0} p( y | \dc X=x, w, u_0 ) p( w | e ) p(u_0 | w) \label{debugging:eqn:drop_do}\\
&= \sum_{w,u_0} p( y | \dc X=x, w, u_0 ) p( w | e ) p(u_0) ,\label{debugging:eqn:approx_cf_rephrased}
\end{align}
where Equation (\ref{debugging:eqn:drop_do}) is due to the fact that the distribution of $U_0$ is invariant and $X$ does not influence $W$, so $W$ can be written as the same function of $U_0$ in $M_0$ and $(M_0)_{\dc X=x}$; and Equation (\ref{debugging:eqn:approx_cf_rephrased}) is due to the fact that $W \subset Z$ and $Z \ind U_0$ by definition of $U_0$.

On the other hand, we have
\begin{align}
&p( Y_{\dc X=x}=y | e) \\
&= \sum_{u: p(u,e) > 0} p( y | \dc X=x, u) p(u | e) \\
&= \sum_{u_0, u_1: p(u_0, u_1,e) > 0} p( y | \dc X=x, u_0, u_1) p(u_0, u_1 | e) \\
&= \sum_{u_0, u_1, w: p(u_0, u_1,e) > 0} p( y, w | \dc X=x, u_0, u_1)  p(u_0, u_1 | e) \\
&= \sum_{u_0, u_1, w: p(u_0, u_1,e), p(u_0, u_1, w) > 0} p( y, w | \dc X=x, u_0, u_1)  p(u_0, u_1 | e) \\
&= \sum_{u_0, u_1, w: p(u_0, u_1,e), p(u_0, u_1, w) > 0} p( y | \dc X=x, u_0, u_1, w) p(w | \dc X=x, u_1, u_0) p(u_0, u_1 | e) \\
&= \sum_{u_0, u_1, w: p(u_0, u_1,e), p(u_0, u_1, w) > 0} p( y | \dc X=x, u_0, w) p(w | \dc X=x, u_1, u_0) p(u_0, u_1 | e) \label{debugging:eqn:throw0} \\
&= \sum_{u_0, u_1, w: p(u_0, u_1,e), p(u_0, u_1, w) > 0} p( y | \dc X=x, u_0, w) p(w | u_1, u_0) p(u_0, u_1 | e) \label{debugging:eqn:throw} \\
&= \sum_{u_0, u_1, w: p(u_0, u_1,e), p(u_0, u_1, w) > 0} p( y | \dc X=x, u_0, w) p(w | u_0, u_1, e) p(u_0, u_1 | e) \label{debugging:eqn:det1}  \\
&= \sum_{u_0, u_1, w: p(u_0, u_1,e), p(u_0, u_1, w) > 0} p( y | \dc X=x, u_0, w) p(w, u_0, u_1 | e)   \\
&= \sum_{u_0, u_1, w: p(u_0, u_1,e,w)  > 0} p( y | \dc X=x, u_0, w) p(w, u_0, u_1 | e)   \\
&= \sum_{u_0, w:p(u_0, e,w) > 0} p( y | \dc X=x, u_0, w) \sum_{u_1:p(u_0, u_1,e,w) > 0} p(w, u_0, u_1 | e)   \\
&= \sum_{u_0, w:p(u_0, e,w) > 0} p( y | \dc X=x, u_0, w) p(w, u_0 | e) \\
&= \sum_{u_0, w} p( y | \dc X=x, u_0, w) p(w, u_0 | e) \label{debugging:eqn:classical_cf_rephrased},
\end{align}
where Equation (\ref{debugging:eqn:throw0}) is due to Markovianity and $(Y \ind An(Z) | Z)_{M}$, which implies $(Y \ind U_1 | Z)_{M}$, and thus $(Y \ind U_1 | W)_{M_{\dc X=x}}$,
Equation (\ref{debugging:eqn:throw}) follows from the fact that $X$ does not influence $W$,
Equation (\ref{debugging:eqn:det1}) follows from the fact that $U_1$ already determines $W$.

Note that $p( w | e ) p(u_0) = 0$ implies $p( w, u_0 | e )=0$ and therefore \[\kl[ p( Y_{\dc X=x} | E ) \| p^W( Y_{\dc X=x} | D ) ] \] is defined.

Now we can calculate
\begin{align}
&\kl[ p( Y_{\dc X=x} | E ) \| p^Z( Y_{\dc X=x} | E ) ]  \\
&= \sum_e p(e)  \kl[ p( Y_{\dc X=x} | e ) \| p^Z( Y_{\dc X=x} | e ) ] \\
&\leq \sum_e p(e) \kl[ p(W, U_0 | e) \| p( W | e ) p(U_0) ] \label{debugging:eqn:monot} \\
&= \sum_{e, w, u_0} p(w, u_0, e) \log \frac{p(w, u_0 | e)}{p(w|e) p(u_0)} \\
&= \sum_{e, w, u_0} p(w, u_0, e) \log \frac{p(w, u_0, e)}{p(w, e) p(u_0)} \\
&= \mi(W, E : U_0) \\
&\leq \mi(Z, E : U_0) \\
&= \mi(Z : U_0) + \mi(E : U_0 | Z) \label{debugging:eqn:chain_mutual} \\
&= 0 + \se(E|Z) - \se(E|U_0,Z) \label{debugging:eqn:indep_noise} ,
\end{align}
where inequality (\ref{debugging:eqn:monot}) follows from the monotonicity (which follows from the chain rule) of the Kullback-Leibler divergence \citep{Cover} together with equations (\ref{debugging:eqn:classical_cf_rephrased}) and (\ref{debugging:eqn:approx_cf_rephrased}),
Equation (\ref{debugging:eqn:chain_mutual}) is the chain rule for mutual information,
and $\mi(W : U_0) = 0$ is due to $U_0$ not influencing $W$ and Markovianity.

\end{proof}

%
%
%

Note that, if we chose the set $Z$ in the above proposition such that it is as ``close'' (in the causal diagram) to $Y$ as possible, this could yield better approximations $p^Z( Y_{\dc X=x} = y| e )$ than simply letting $Z$ be the root nodes, as done in $\bar{p}( Y_{\dc X=x} = y| e )$.
We leave this as a question for future work.

\subsection{Proof of Proposition \ref{debugging:prop:approx_transf}}
\label{debugging:sec:prop_transf}

Here we give a proof for Proposition \ref{debugging:prop:approx_transf}.


\begin{proof}
We calculate
\begin{align}
&\kl( p(Z) \| \bar{p}(Z) ) \label{debugging:eqn:mo} \\
&\leq \kl( p(X_0,\ldots,X_K,C) \| p(C) \Pi_k p(X_k|C) ) \notag \\
&= \kl( p(C) p(X_0|C) p(X_1|X_0,C) \cdots p(X_K|X_0,\ldots,X_{K-1},C) \| p(X) \Pi_k p(X_k|C) ) \notag \\
&= \sum_{x_0,\ldots,x_K,c} p(x_0,\ldots,x_K,c) \log \frac{p(c)}{p(c)} \frac{p(x_0|c)}{p(x_0|c)} \frac{p(x_1|x_0,c)}{p(x_1|c)} \frac{p(x_2|x_0,x_1,c)}{p(x_2|c)} \cdots \notag\\
&\phantom{=} \cdot \frac{p(x_K|x_0,\ldots,x_{K-1},c)}{p(x_K|c)} \notag \\
&= \sum_{x_0,\ldots,x_K,c} p(x_0,\ldots,x_K,c) \log \frac{p(c)}{p(c)} \frac{p(x_0|c)}{p(x_0|c)} \frac{p(x_1,x_0|c)}{p(x_1|c)p(x_0|c)} \frac{p(x_2,x_0,x_1|c)}{p(x_2|c)p(x_0,x_1|c)} \cdots \notag \\
&\phantom{=} \cdot \frac{p(x_K,x_0,\ldots,x_{K-1}|c)}{p(x_K|c)p(x_0,\ldots,x_{K-1}|c)} \notag \\
&=\mi(X_1 : X_0 | C) + \mi(X_2 : X_0, X_1 | C) + \ldots + \mi(X_K : X_0, \ldots, X_{K-1} |C ) \notag \\
&\leq\se(X_1 | C) + \se(X_2 | C) + \ldots + \se(X_K |C )  \notag , \\
\end{align}
where inequality (\ref{debugging:eqn:mo}) follows from the monotonicity (which follows from the chain rule) of the Kullback-Leibler divergence \citep{Cover}.
\end{proof}

\bibliographystyle{abbrvnat}
\bibliography{include/causal_debugging,include/phil_master,include/diss}

\end{document}